\documentclass[a4paper,twoside]{article}

\usepackage{epsfig}
\usepackage{subcaption}
\usepackage{calc}
\usepackage{amssymb}
\usepackage{amstext}
\usepackage{amsmath}
\usepackage{amsthm}
\usepackage{multicol}
\usepackage{pslatex}
\usepackage{apalike}
\usepackage[bottom]{footmisc}
\usepackage{xcolor}
\usepackage{multirow}
\everypar{\looseness=-1}

\usepackage{SCITEPRESS}     

\begin{document}

\title{Predicting Eye Gaze Location on Websites}
\author{\authorname{Ciheng Zhang\sup{1}, Decky Aspandi\sup{2}, Steffen Staab\sup{2}\sup{3}} 
\affiliation{\sup{1}Institute of Industrial Automation and Software Engineering,  University of Stuttgart, Stuttgart, Germany}
\affiliation{\sup{2}Institute for Parallel and Distributed Systems,  University of Stuttgart,  Stuttgart, Germany}
\affiliation{\sup{3}Web and Internet Science, University of Southampton, Southampton, United Kingdom}
\email{zhang.ciheng@dotted-eighth.com,\{decky.aspandi-latif, steffen.staab\}@ipvs.uni-stuttgart.de}
}


\abstract{World-wide-web, with website and webpage as a main interface, facilitates dissemination of important information. Hence it is crucial to optimize webpage design for better user interaction, which is primarily done by analyzing users' behavior, especially users' eye-gaze locations on the webpage. However, gathering these data is still considered to be labor and time intensive. In this work, we enable the development of automatic eye-gaze estimations given webpage screenshots as input by curating of a unified dataset that consists of webpage screenshots, eye-gaze heatmap and website's layout information in the form of image and text masks. Our curated dataset allows us to propose a deep learning-based model that leverages on both webpage screenshot and content information (image and text spatial location), which are then combined through attention mechanism for effective eye-gaze prediction. In our experiment, we show benefits of careful fine-tuning using our unified dataset to improve accuracy of eye-gaze predictions. We further observe the capability of our model to focus on targeted areas (images and text) to achieve accurate eye-gaze area predictions. Finally, comparison with other alternatives shows state-of-the-art result of our approach, establishing a benchmark for webpage based eye-gaze prediction task. }






\onecolumn \maketitle \normalsize \setcounter{footnote}{0} \vfill

\section{\uppercase{Introduction}}
\label{sec:introduction}

%



Analysis of user behavior during their interaction with the website (and contained webpages) is important to evaluate the website overall quality. This information can be used to create more optimized webpages for better interactions, as such, there are needs to characterize these behaviors. Some of the characteristics include users' tendency to focus on certain areas (e.g. upper left corner) during browsing~\cite{shen2014webpage}. Other important characteristics can be defined from users' eye-gaze data, which normally represents visual attention of the users during website interaction. This data is usually acquired from human pupil-retinal location and movement relative to the object of interest that allows one to pinpoint exact webpage area, where users focus during their interactions and its relations with overall webpage structure. Although this information is crucial to create a more optimized website for better interactions, However, acquiring these eye-gaze data from every user during their browsing duration is difficult, given the complexity of the acquisition setting. Thus there is currently a need for an automatic approach to predict these eye-gaze locations given an observation of the webpages.

In computer vision field, saliency prediction task has been extensively investigated that allows estimation of people's attention given an image. The main task is to estimate saliency map, which highlights first location (or region) where observers' eyes focus, with photographs of natural scenes~\cite{zhang2018review,kroner2020contextual} commonly used as input. Several methods have been developed so far to solve this task, including machine learning~\cite{hou2008dynamic,li2012visual} and recently deep learning based approaches~\cite{kroner2020contextual} with quite high accuracy achieved. With this progress, there is an opportunity to adopt these automatic, visual based saliency predictors for eye-gaze predictions task, by representing input observation in a form of webpage screenshot. This enables adaptation of task objective for eye-gaze heatmap prediction, in lieu of saliency map, with both predicted maps representing area where most people (or in our case users) attend.

The challenges however are differences between saliency and eye-gaze prediction tasks in terms of content, structure and layout of input images. For instance, there is not any concept of depth on webpage screenshot as opposed to natural images, an aspect that is highly utilized for general visual saliency estimation. In addition, high contrast and varied colored areas on natural image are commonly regarded as saliency area, which may not be the case for webpage screenshots, given that user eye-gaze locations are highly dependent on the type of interactions during website browsing. This problem can be rectified by the use of a data-driven approach~\cite{5288526}, which can be made feasible through fine-tuning using a specialized dataset on a particular task. However, this attempt is currently hampered by the lack of such dataset. 

In this work, we curate a generalized dataset of eye-gaze prediction from webpage screenshots to enable effective training for machine and deep learning approaches. Using this dataset, we propose a deep learning based method that incorporates image and textual locations of webpage through mask modalities and combines them with attention fusion mechanism. We then evaluate the impact of transfer learning, including comparison with other alternatives, to establish a benchmark for this task. Specifically, the contributions of this work are: 

\begin{enumerate} 
    \item The establishment of a unified benchmark dataset for eye-gaze detection from webpage screenshots, derived from website and user interactions data. 
    \item A novel deep learning-based and multi-modal eye-gaze detector with internal attention that leverages characteristics of input contents and importance of each stream of modality for accurate eye-gaze predictions.
    \item Benchmark results of automatic eye-gaze location estimators and our state-of-the-art results for eye-gaze detection task given webpage screenshot inputs.
\end{enumerate}

\section{\uppercase{Related Work}}
An early example of work analyzing website content is done by Shen and Zhao et.al.~\cite{shen2014webpage} where three types of webpages are analyzed: Text-based, Pictorial-based, and Mixed (combination of Text and Pictorial) websites. The authors show that some attention characteristics of users during their interactions with webpage exist, with main finding that users usually pay more attention to some relevant parts of websites (such as left-top corner of a website) and their tendency to focus on areas where large images are present. Furthermore, on the websites from ‘Text’ category, their preference to focus on certain parts of websites (middle left and bottom left regions) is perceived. Lastly, they propose multi-kernel machine learning inferences for eye-gaze heatmap predictions (which represent visual attention of users) for their developed website eye-gaze dataset of Fixations in Webpage Images dataset (FiWI).

In computer vision field, the task of locating (or predicting) users' attention to natural image is commonly called as saliency prediction. These tasks are commonly solved by utilising machine learning techniques, given their automation capability. An earliest example of this method is Incremental Coding Length (ICL) ~\cite{hou2008dynamic} which aims to predict the activation of saliency location by measuring perspective entropy gain of each input feature (several image patches) as a linear combination of sparse coding basis function. Other algorithm of Context-Aware Saliency Detection (CASD) capitalises on the concept of dominant objects as additional context to improve their prediction~\cite{goferman2011context}. Furthermore, D Houx et.al. ~\cite{houx2012image} propose an approach that aims to solve figure-ground separation problem for prediction. They use a binary and holistic image descriptor of Image Signature, which is defined as sign function of Discrete Cosine Transform (DCT) of an image as additional input. Subsequently, Hypercomplex Fourier Transform (HFT)~\cite{li2012visual} is used to transform input image to frequency domain to estimate saliency map. One relevant work utilizes deep learning based method, given their accurate estimations and ability to leverage huge datasets size (that are increasingly present). This approach is based on Encoder and Decoder structure with a pre-trained VGG network to predict saliency maps~\cite{kroner2020contextual}. Finally in recent years, several researchers extend the saliency approach to work with image and depth input. One example is the work from Zhou et. al. ~\cite{9320595} that uses Hierarchical Multimodal Fusion Network to process both RGB (Red, Gren and Blue) colours image and depth maps (one channel) as additional modality for the input, which results in more accurate gaze maps prediction.

Even though all of the described methods work for general visual saliency predictions, however, their capability to predict users' eye-gaze location on webpage is not yet investigated, given lack of dataset available. Thus in this work, we propose to create a specialized webpage based eye-gaze prediction datasets to allow for the development of automatic eye-gaze predictions of webpage screenshot inputs, then utilize it to develop our deep learning based eye-gaze location predictor.




\begin{figure*}[h!]
    \centering
    \includegraphics[width=0.8\linewidth]{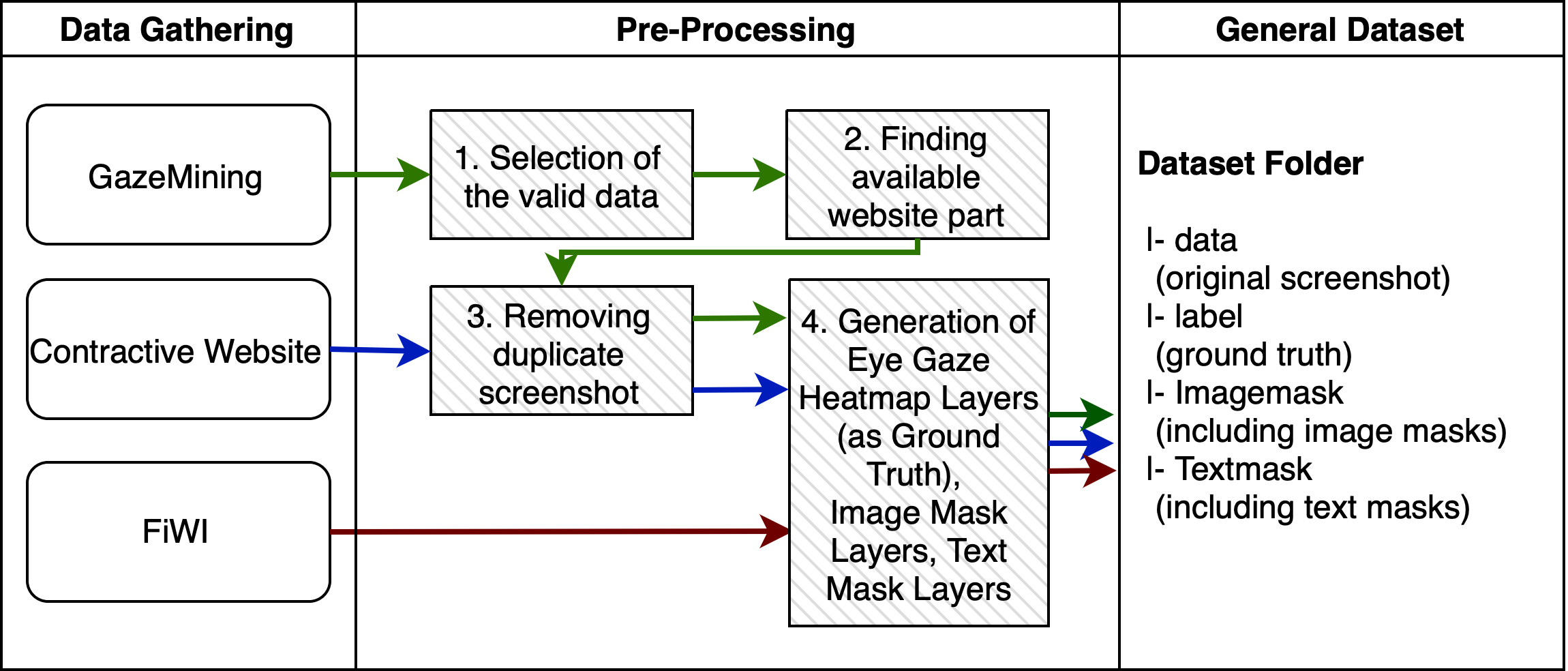}
    \caption{The flow of dataset generation.}
    \label{fig:datsáset}
\end{figure*}

\section{\uppercase{Methodology}}
\label{sec:meth}

\subsection{Dataset Gathering and Processing}
\label{sec:dataset}
In this work, we search, process and curate a generalized eye-gaze dataset from available datasets in literature. Here, we focus on three publicly available user-website interaction datasets (where eye-gaze and webpage screenshots are present): GazeMining~\cite{menges_2020}, Contrastive Website~\cite{aspandi2022user}, and Fixations in Webpage Images (FiWI) dataset~\cite{shen2014webpage}. In this section, we provide a short description of datasets, and proceed with description of pre-processing algorithm that we conduct to produce our unified dataset. The processed dataset is available at \footnote{https://doi.org/10.18419/darus-3251} and available after request.


\begin{itemize}
    \item \textbf{GazeMining} is a dataset of video and interaction recordings on dynamic webpages ~\cite{menges_2020}. In this work, the authors focused on websites which contain constantly changing scenes (thus dynamic). The data was collected on March 2019 from four participants who interacted with 12 different websites. 
    \item \textbf{Contrastive Website dataset} is a recent website interaction-based dataset, where participants were asked to visit two sets of websites (each of four) and performed respective tasks (flight planning, route search, new searching and online shopping) resulting on more than 160 sessions~\cite{aspandi2022user}.  
    \item \textbf{FiWI} dataset~\cite{shen2014webpage} is a dataset that focuses on website-based saliency prediction. However, it is small in term of webpage screenshot availability (only 149 images) compared to two previous datasets that prevents its use for larger scale evaluations (this is especially true for a deep learning model). Therefore FiWI dataset is mostly (and commonly) utilized as comparative evaluation with other models, but not for model training.
\end{itemize}

\begin{figure*}[h!]
    \centering
    \includegraphics[width=0.95\linewidth]{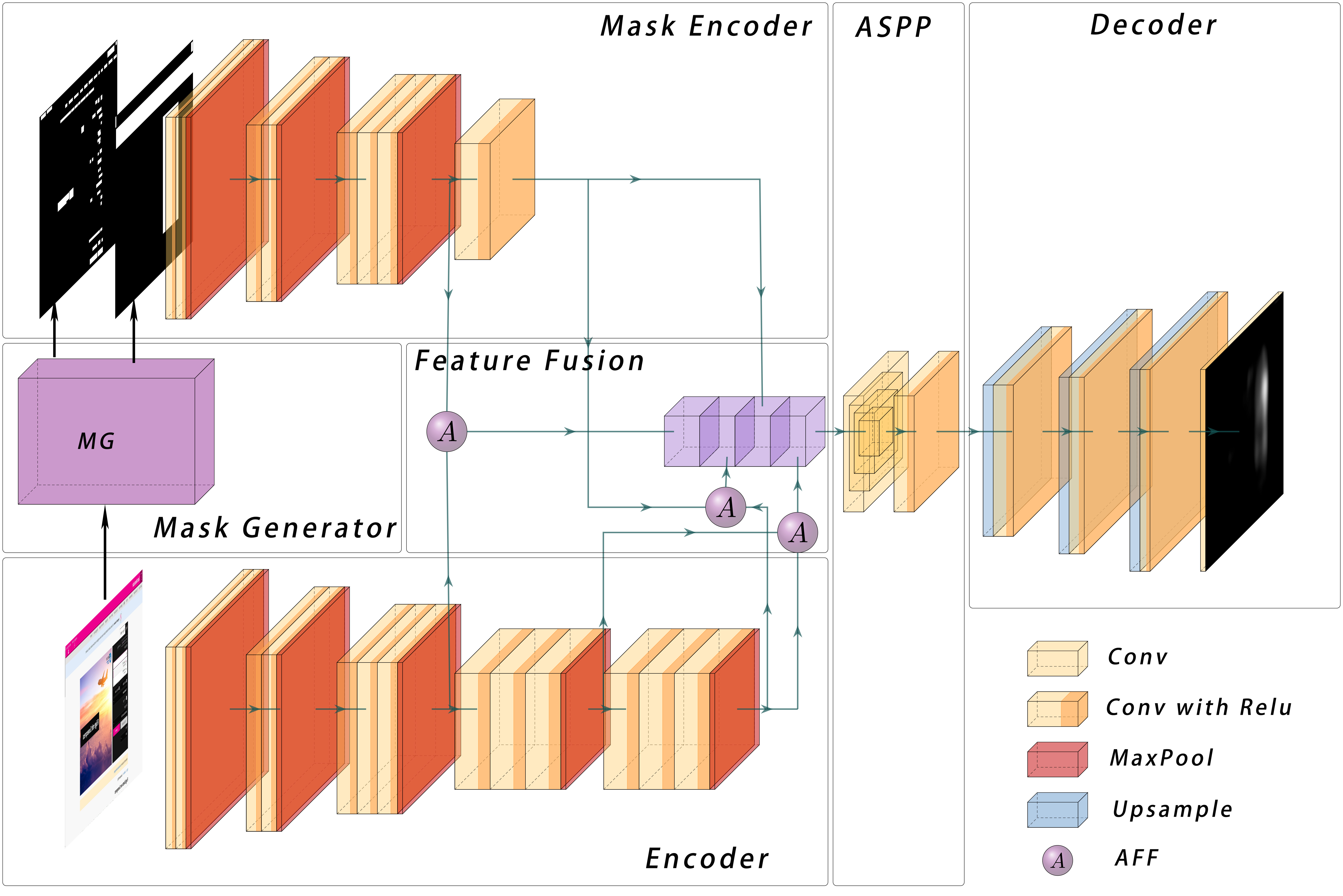}
    \caption{Structure of Multi Mask Input Attentional Network.}
    \label{fig:basicmodell}
\end{figure*}


Figure~\ref{fig:datsáset} shows the flow of our dataset curation, data pre-processing and the results of the generalized dataset. The first step of pre-processing is elimination of empty screenshots (which exist in small quantities on GazeMining dataset), i.e. no screenshots are present, which can come due to rapid sampling during acquisition. Secondly, due to dynamic nature of the observed websites, webpages' static parts (that consist of all unchanged, rendered webpages' elements during interaction time) are removed (or blackened) on original datasets, which necessitates us to find and collect only recorded screenshot - thus removing irrelevant observations (black parts). Thirdly, we remove the duplicate webpage screenshots of both datasets for efficiency and generate respective locations of Image and Text as independent layers (called Image and Text Mask, which we will detail in Section~\ref{sec:inputMask}). Lastly, eye-gaze heatmap layers are generated (as ground truth) with respect to duration of each observed eye-gaze. This pipeline is applied to all datasets, with the exception of Contrastive Website, where first and second steps are skipped, and for FiWI where only last step is necessary. In the end, our unified dataset includes four sub-folders (data, label, Imagemask, TextMask) with a total of 3119 screenshot examples  (1546 samples or 49.6\% from GazeMining, 1424 instances or 45.7\% from Contrastive Website dataset and 149 examples or 4.7\% from FiWI) along with the associated ground truths.


    


\subsection{Multi Mask Input Attentional Network (MMIAN) for Eye Gaze Prediction}
Our pre-processed datasets allows us to propose a deep learning based model to benefit from sizable number of observations~\cite{rodriguez2021machine}. Here we propose a Multi Mask Input Attentional Network (MMIAN) that aims to predict eye gaze location given a webpage screenshot as input. Specifically, given input webpage screenshot images of $X_{i..n}$, with $X$ as 2D matrix of webpage screenshot image, and n as number of batch size, MMIAN estimates eye-gaze locations of $\hat{Y}_{i..n}$, where each $\hat{Y}$ is a 2D matrix of eye-gaze heatmap, with each cell value represents probability of eye-gaze locations. The structure of MMIAN follows encoder-decoder scheme that is inspired by Multi-scale Information Network (MSI-Model) ~\cite{kroner2020contextual} consisting of several modules: input and Mask Encoder, Mask Generator, Attentional Feature Fusion, Atrous Spatial Pyramid Pooling (ASPP) Module and Decoder. In this work, we further propose several important modifications: 

\begin{enumerate}
    \item Incorporation of several input masks to include webpage content information of textual and images spatial location. 
    \item Addition of attention mechanisms for effective fusion of input modalities.
    
\end{enumerate}

Overall architecture of MMIAN can be seen in Figure~\ref{fig:basicmodell}. Specifically, an input webpage screenshot image is passed into mask generator to produce an image mask and text mask. Then both masks are concatenated and fed into Mask Encoder to extract relevant features, while input image is simultaneously fed into input Encoder. Feature Fusion module then fuses extracted mask features and input features with several attentional modules and combines them in a conjoint block. This feature block is then fed into ASPP part to enlarge field of views, capturing different resolution views from the input, thus producing richer feature representations. These features are then passed through series of up-sampling layers (Decoder) creating an eye-gaze saliency map as a final result. The implementation code of our model is available on our repository\footnote{https://github.com/ZackCHZhang/WebToGaze}.





\subsubsection{Input Masks Generator}
\label{sec:inputMask}

\begin{figure*}[h]
\centering

 \subfloat[Input]{
 \includegraphics[width=0.33\linewidth]{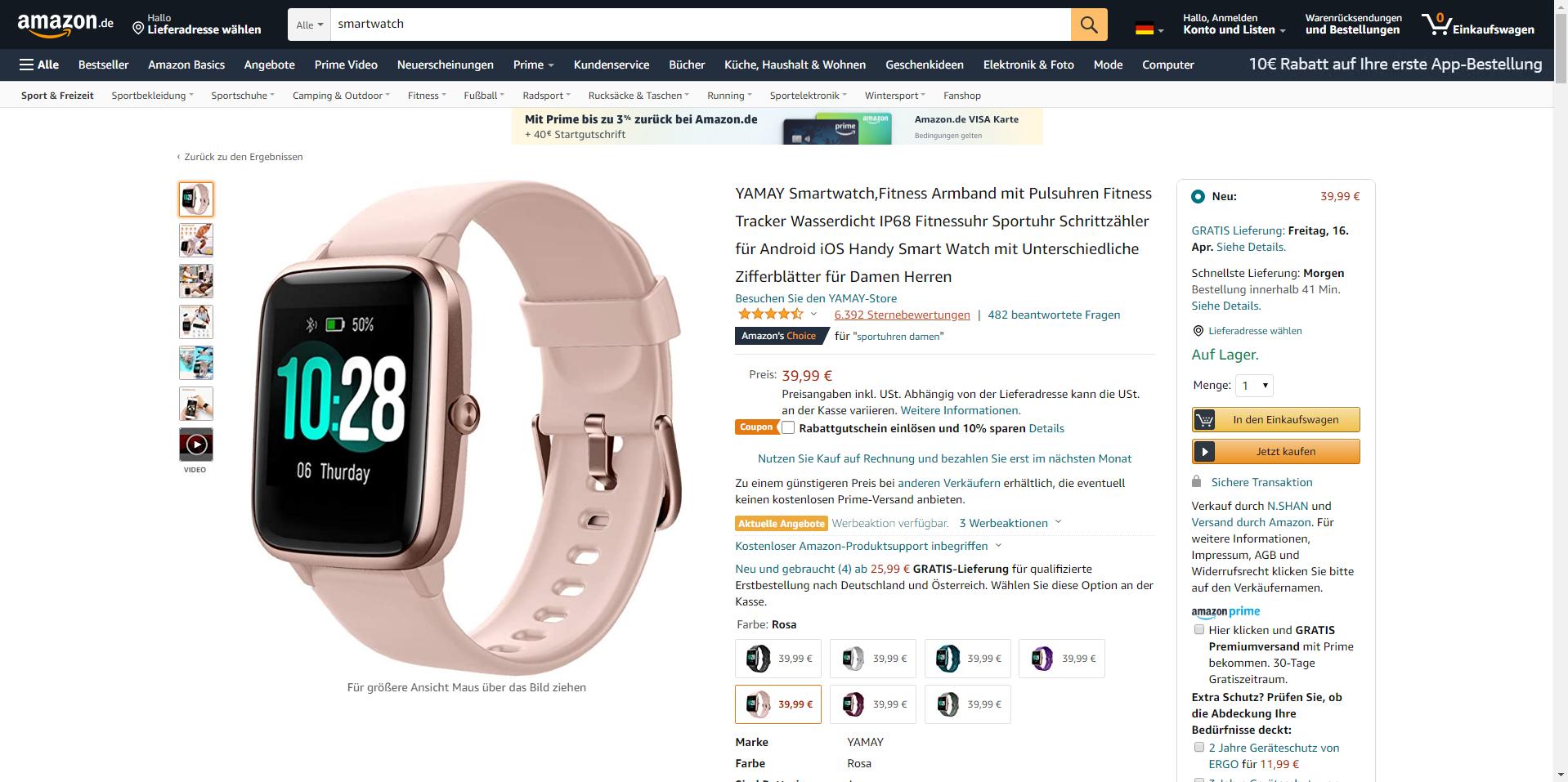}} 
 \hfill
 \subfloat[Text Mask]{\includegraphics[width=0.33\linewidth]{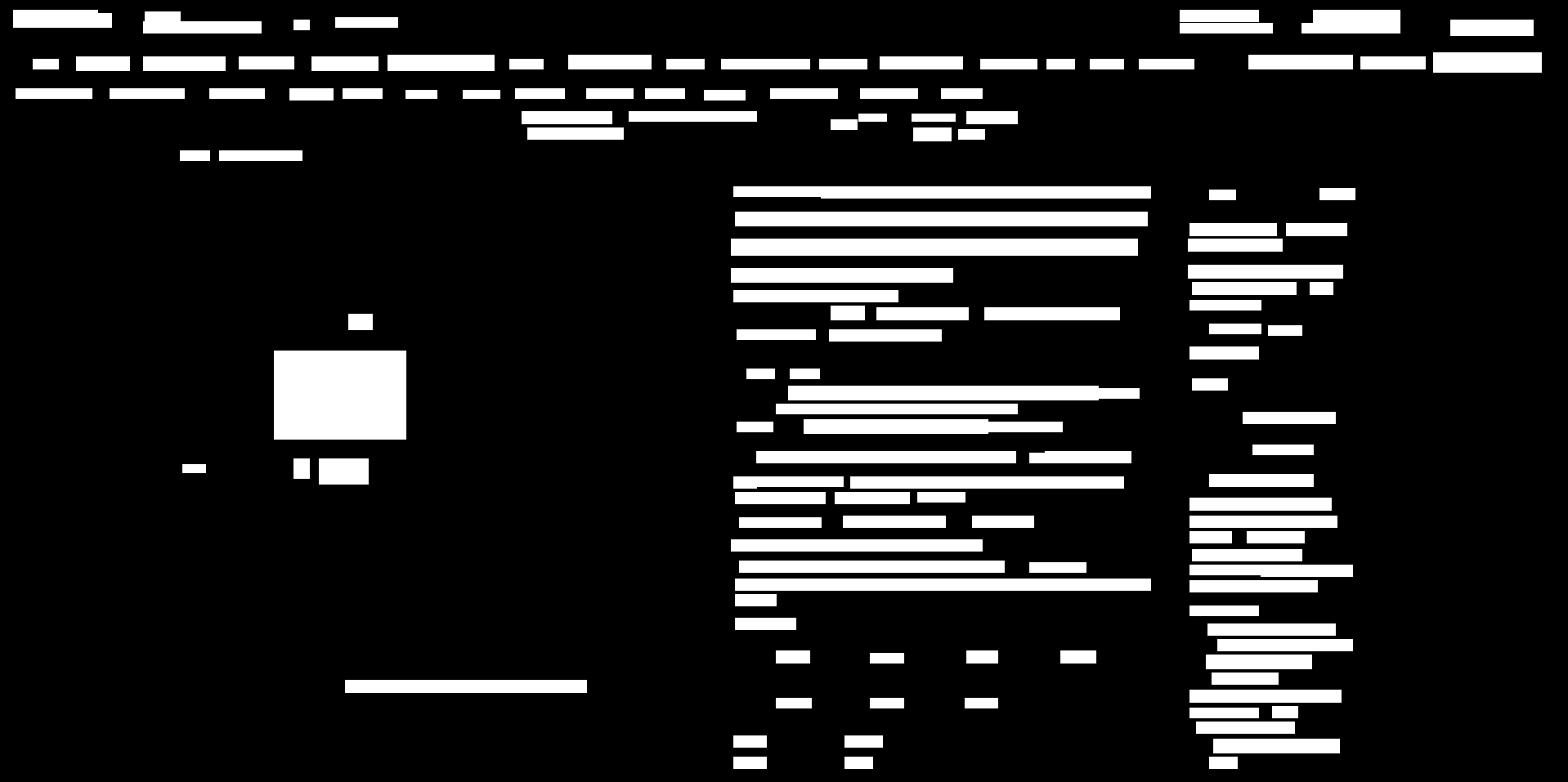}}
 \hfill
 \subfloat[Image Mask]{\includegraphics[width=0.33\linewidth]{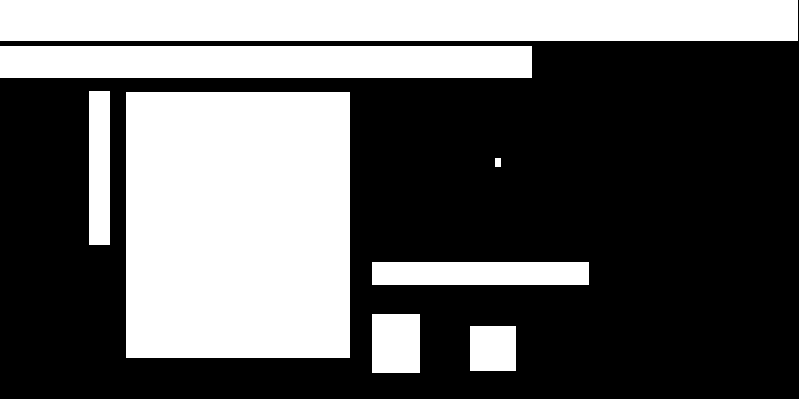}}

\caption{One case of input and text mask generation. Figure (a) is an example of webpage screenshot, Figure (b) and (c) are the generated text mask and image mask respectively.}
\label{fig:mask}
\end{figure*}

\label{subsubsec:maskG}
It has been observed that webpage layout (which in principle is arrangement of images and texts) is an important aspect of webpage~\cite{shen2014webpage}. To benefit from this information, we propose to incorporate spatial locations of both image and text on websites with mask representations: image and text masks. These masks are represented in form of a matrix, where each cell is activated in the presence of both image and text in webpage screenshot.  
\begin{itemize}
    \item \textbf{Image Mask Generation: } For image location recognition, we use method from M. Xie et al.~\cite{10.1145/3368089.3417940} which produces bounding-box locations of an input image in a binary map (i.e. map value is set to 1 where image is present, otherwise 0). 
    
 
  \item  \textbf{Text Mask Generation: }we use one of available optical character recognition (OCR) methods of Efficient and Accuracy Scene Text detection (EAST)~\cite{Zhou_2017_CVPR} to locate location of text of webpage screenshot input. Similar to Image Mask, this process generates corresponding Text Mask in Binary map format.
  

\end{itemize}



One example of generated masks from our curated dataset can be seen in Figure~\ref{fig:mask}. We can see that both of masks contain quite accurate locations of both Image and Text on an input webpage screenshot. Even though some mild imperfections occur (i.e. some buttons can be recognized as images, it does not fully detect locations of certain images due to small size, and some text within images are included on text masks), however in general the locations of both image and text on input webpage screenshots are properly recognized. These masks are then concatenated to be fed into Mask Encoder part.



\subsubsection{Input and Mask Encoder}
Encoder part is a series of Convolutional Neural Network that aims to extract visual features from input matrix, which is original web screenshot image and concatenated generated Masks, for Input and Mask Encoder respectively. Both Encoders are based on VGG16~\cite{simonyan2014very} architectures with last fully connected layer removed, and further reductions of convolution layers (five convolutional layers with Relu and two max-polling layers) applied for Mask Encoder. Input and Mask encoder then produce MaskFeatures and Screenshot Features to be combined through Multi-Modal Attentional Fusion module.

\subsubsection{Multi-modal Attentional Fusion}

\begin{figure}[h]
    \centering
    \includegraphics[width=1\linewidth]{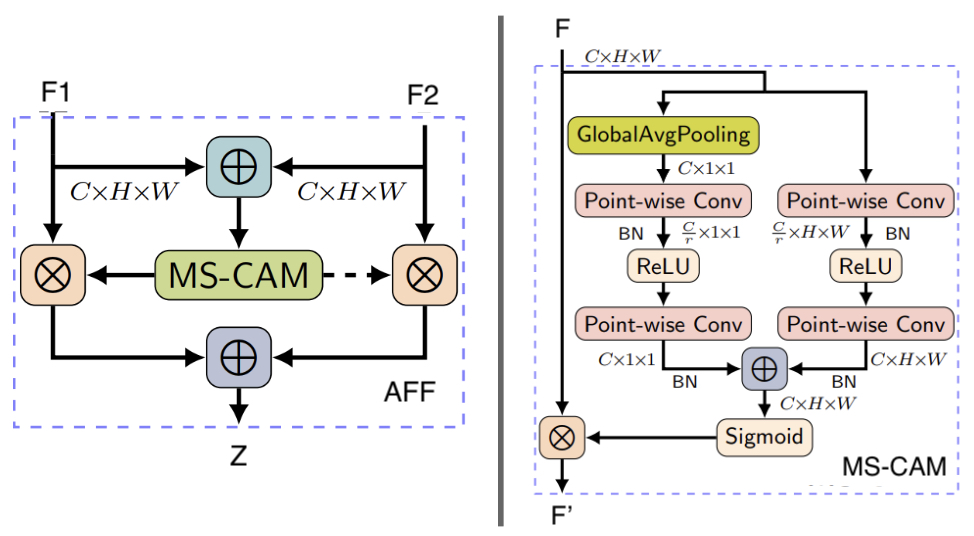}
    \caption{Left is Attentional Feature Fusion block, right is Multi-scale Channel Attention module~\cite{dai2021attentional}.}
    \label{fig:aff}
\end{figure}

We employ multi-modal inputs (that consists of web-page screenshot and generated masks) to benefit from additional stream of information~\cite{9320595,aspandi2020latent}. Moreover, we introduce attentional fusion mechanism for more effective fusion of these modalities, as opposed to simple fusion operations of summation or concatenation~\cite{aspandi2022audio}, and further solving the problems of inconsistent input semantics and scales~\cite{dai2021attentional}. An example of attentional feature fusion block (AFF) can be seen in Figure~\ref{fig:aff} where it receives two streams of inputs (denoted as F1 and F2) and initially fuses these input features through matrix addition (denoted as $\bigoplus$). The result from addition of internal Multi-scale Channel Attention module (that aggregates channel context by a series of point-wise convolutions) and a sigmoid function generates a fusion weight. This fusion weight and its another counterpart (where it is negated by one) are multiplied (denoted as $\bigotimes$) by each of original inputs respectively, and then added together ($\bigoplus$ symbol) to produce their weighted average as fused feature of Z. This fusion process is applied to the received MaskFeatures and Screenshot Features three times, to allow for observation of different scales of respective features (these operations are marked as circled A in Figure~\ref{fig:basicmodell}). The resultants of the fused features are then concatenated to a conjoint block for subsequent pipelines.





\subsubsection{Atrous Spatial Pyramid Pooling Module (ASPP)}
ASPP module is made by superimposing different atrous convolutions~\cite{chen2017deeplab} filters to obtain features from larger receptive fields. This is beneficial given that it is common for webpage layouts to be modularized into a variety of different sub-layouts. This thus increases receptive fields (or field-of-view) of internal convolutional layers (where the result from several filters are combined together) enables us to capture such layout configurations. 
In our ASPP module, we use dilation rates of 4, 8, and 12, with number of $1\times1$ filter to be 256 to ensure kernel size compatibility. This module is applied to conjoint block input, and subsequent features are obtained for decoding operation.





\subsubsection{Output Decoder}
Decoder part consists of a series of convolution layers with up-sampling to decode input features, generating final prediction of eye-gaze heatmap location (in the form of a 2D tensor and in binary mode). Specifically, number of up-sampling blocks consisting of bilinear scaling operations (each sequentially doubles the number of input tensor) and a subsequent convolutional layer with kernel size $3 \times 3$ are used. With this module, we provide the features generated from ASPP as input features and final eye-gaze heatmap predictions are obtained. Finally, we convert binary heatmap to gray-scale format, to conform to the original scale of eye-gaze heatmap from ground-truth map.

\subsection{Training Procedure}
To evaluate the impact of training of models utilizing webpage screenshot based eye-gaze prediction datasets, first we perform training process using an original saliency prediction model of Kroner's Model (MSI-Model)~\cite{kroner2020contextual}. This model has been pre-trained on general visual saliency dataset (SALICON)~\cite{jiang2015salicon}, which we call as P-MSI, that serves as baseline. Given this pre-trained model, we perform fine-tuning using our pre-processed datasets (GazeMining and Contractive Website dataset) producing FT-MSI-GM and FT-MSI-CW respectively, and evaluate observed accuracy gains. We did the model fine-tuning instead of performing training from scratch to maximize the potential accuracy obtained - we empirically found that the overall obtained accuracy of the latter to be substantially lower.

Additionally, we propose to evaluate the impact of training using a combined dataset (i.e. we combine examples of both of GazeMining and Contrastive Website) by further training MSI-Model with this merged dataset, that we name as FT-MSI-CMB. The training is conducted until convergence and no further accuracy improvement is perceived. Afterward, we proceed to training stage of our proposed model of MMIAN by first transferring Encoder and Decoder's weight from best performing models of the previous step, and initialized weights of both Mask Encoder and ASPP by zero-mean uniform distribution.

All of the training is conducted by minimizing Kullback-Leibler divergence (KLD) loss function as shown in Equation~\ref{eq:kld}, with $Y$ indicating ground truth and $\epsilon$ is a regularization constant to guarantee that denominator is not zero. In this loss, estimation of saliency maps can be regarded as a probability distribution prediction task, as formulated by Jetley et. al.~\cite{jetley2016end}. The output of estimator is normalized to a non-negative number, with KLD value used to measure level of differences between predictions and ground truth. Finally, Adam optimizer~\cite{kingma2014adam} with a learning rate of $10^{-4}$ is used for overall optimization.  

\begin{equation}
\label{eq:kld}
    D_{KL}(\hat{Y}||Y) = \sum_i~Y_i~ln(\epsilon + \frac{Y_i}{\epsilon + \hat{Y}_i})
\end{equation}

\subsection{Experiment Setting}

\subsubsection{Dataset Experiment Setting}
In this experiment, all of three pre-processed datasets (cf. Section~\ref{sec:dataset}) are utilized, with training, validation and test instance numbers are shown in Table~\ref{tab:dataset}. Specifically, both GazeMining and Contrastive Website datasets are divided with 60\%, 20\% and 20\% of samples for training, validation set, and test set respectively. Whereas for FiWI dataset~\cite{shen2014webpage}, all samples (of 149 images) are used for testing.


\begin{table}[!h]
    \centering
\resizebox{1\linewidth}{!}{
    \begin{tabular}{l|c|c|c}
    \hline
 Splits & GazeMining & Contrastive Website & FiWI \\
 \hline
 Training&  928 & 868 & - \\
 Validation & 309 & 278 & -\\
 Test& 309 & 278 & 149\\
    \hline
    \end{tabular}}
    \caption{Training, validation and test set for each dataset.}
    \label{tab:dataset}
\end{table}

\subsubsection{Quantitative Metrics}
We use three quantitative metrics of Area under Receiver Operating Characteristic curve (AUC), Normalized Scanpath Saliency (NSS), and Person's Correlation Coefficient (CC) to judge the quality of models' prediction.


\begin{figure*}[h]
\centering

 \subfloat[Input+GT]{
 \includegraphics[width=0.33\textwidth]{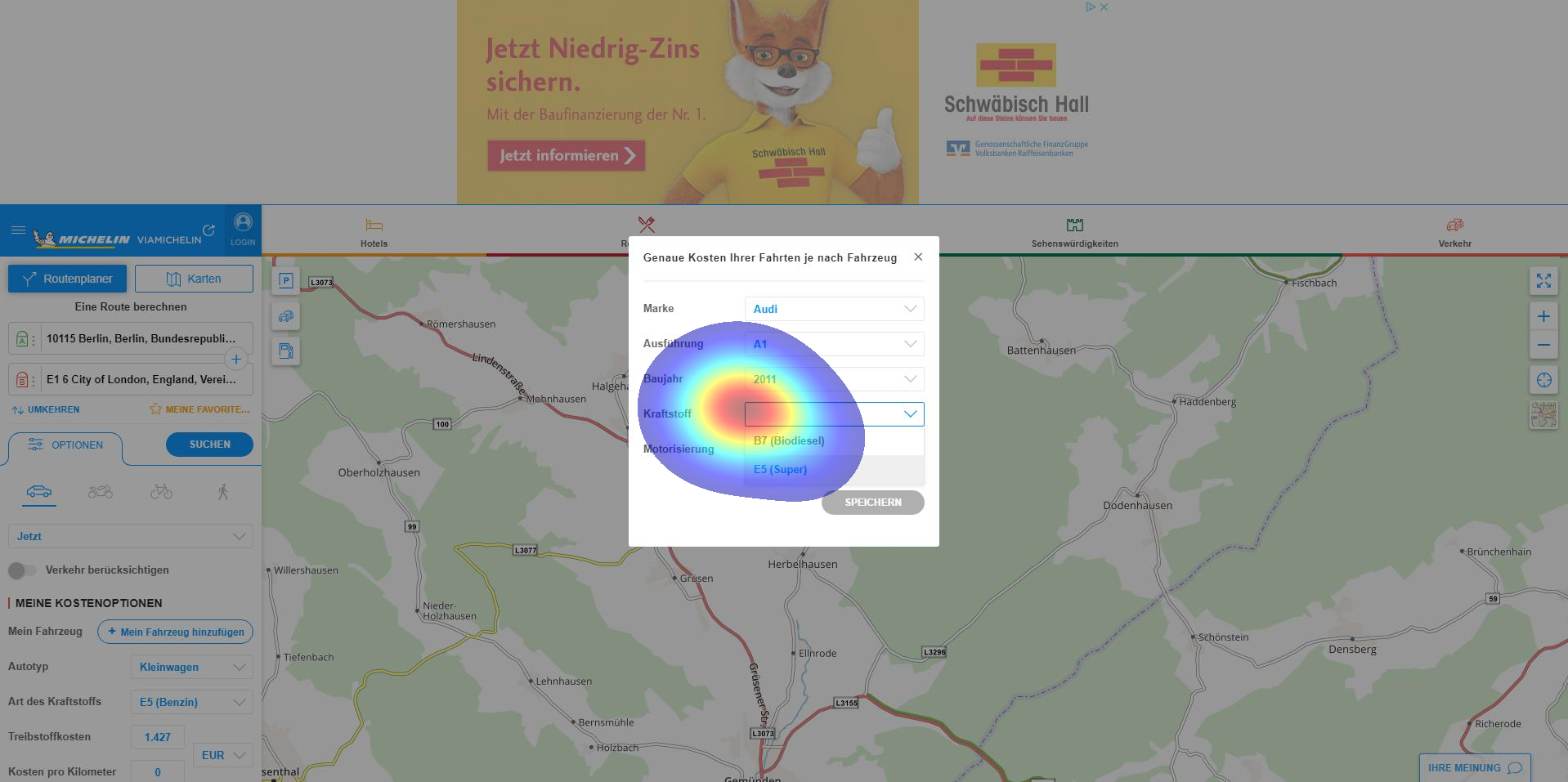}}
 \hfill
 \subfloat[P-MSI]{\includegraphics[width=0.33\textwidth]{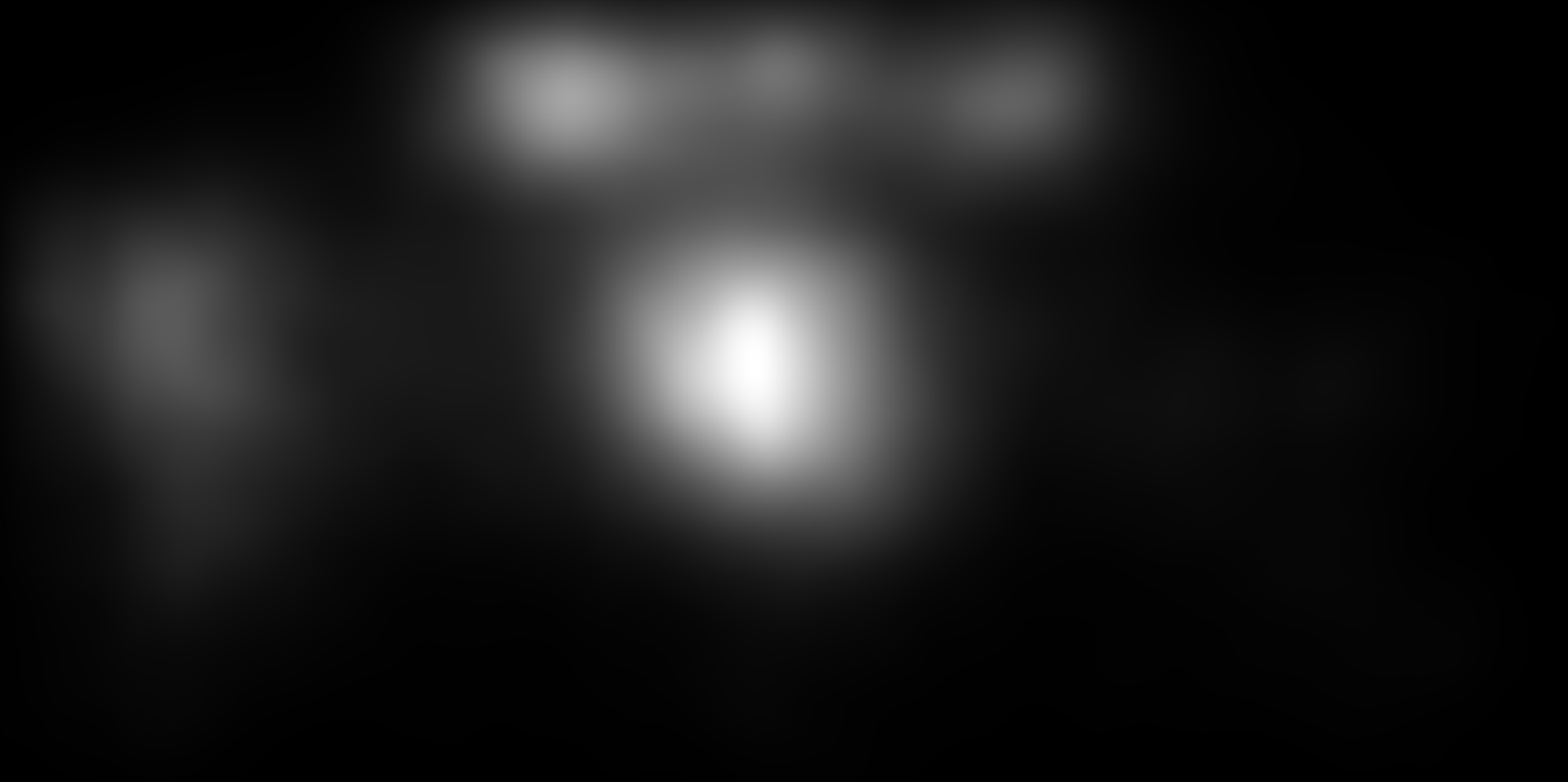}}
 \hfill
 \subfloat[FT-MSI-CW]{\includegraphics[width=0.33\textwidth]{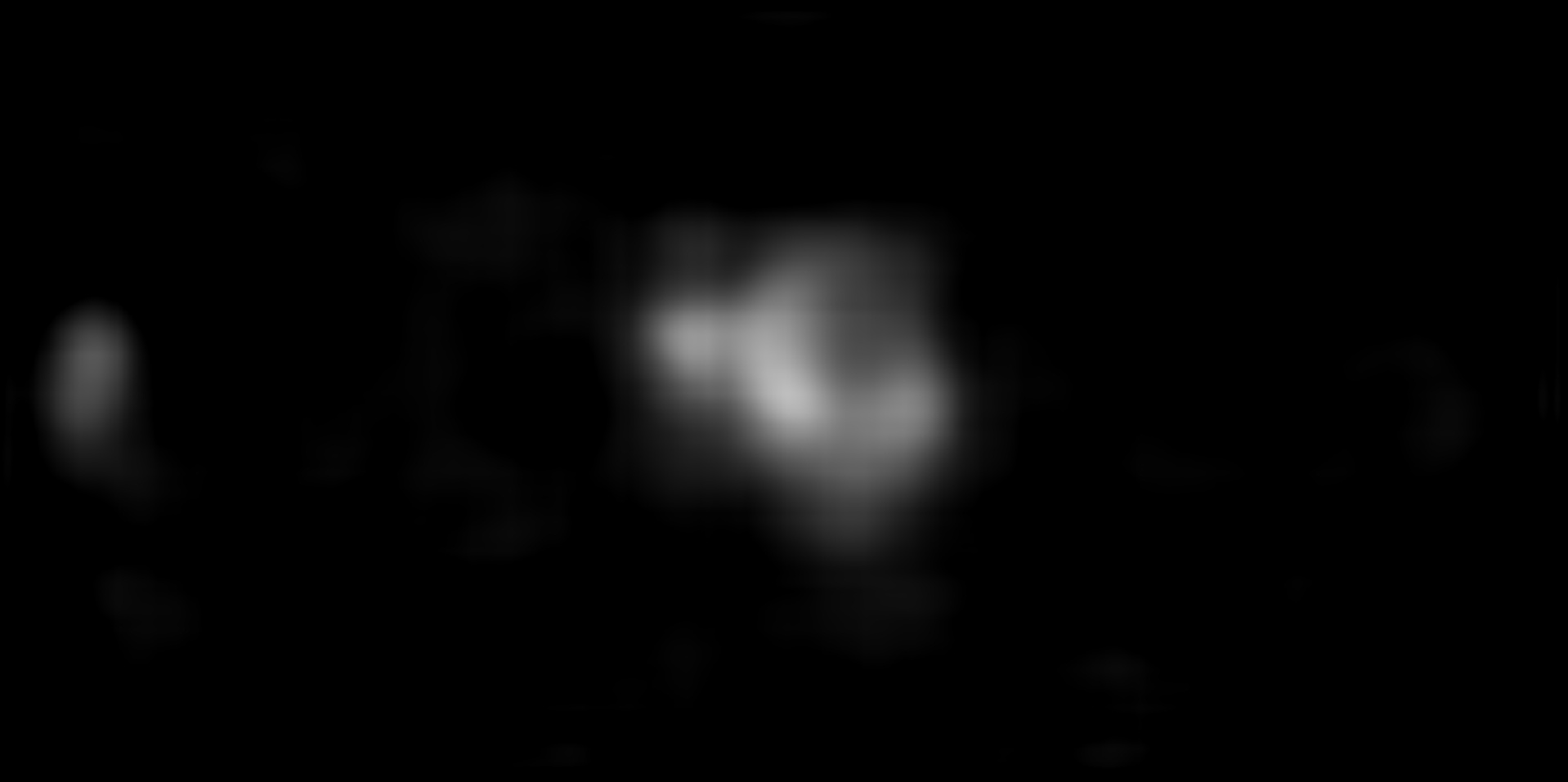}}
 
\caption{Comparison between pre-trained results from P-MSI (baseline) and FT-MSI-CW (Fine-tuned). Figure (a) shows webpage screenshot with ground truth (eye-gaze heatmap). Figure (b) and (c) show predictions from P-MSI and FT-MSI-CW.}
\label{fig:comFT}
\end{figure*}


\begin{itemize}

    \item \textbf{NSS} is commonly used for general saliency prediction tasks as a direct correspondence measure between predicted saliency maps and ground truth, which is computed as average normalized saliency at fixated locations~\cite{PETERS20052397}. Furthermore, NSS is sensitive to false positives, relative differences in saliency across evaluated image, and general monotonic transformations~\cite{bylinskii2018different}. With $Y^B$ as a binary map of true fixation location and $N$ to indicate total pixel number and $i$ indicates each pixel instance, NSS value can be calculated using an equation as shown below:
    \begin{equation}
    \label{eq:nss}
        NSS(\hat{Y},Y^B) = \frac{1}{N}\sum_i \hat{Y}_i \times Y_i^B
    \end{equation}
    
    \item \textbf{AUC-J} evaluates predicted eye-gaze heatmaps as a classification task, where each prediction pixel is evaluated through a binary classification setting. Here, a certain threshold value is used to decide whether it is deemed to be correctly predicted as eye-gaze locations, thus emphasizing frequency of true positive. We use method described by Judd et. al.~\cite{judd2009learning} to select all required thresholds. 

    \item \textbf{CC} metric aims to evaluate level of linear relationship between two input distributions. In eye-gaze location prediction task, both generated eye-gaze location map and ground truth are treated as random variables~\cite{le2007predicting}, and level value is calculated with both of these map inputs following equation~\ref{eq:cc}. Thus, with operator $\sigma$ as covariance matrix, CC value can be calculated as: 
    \begin{equation}
    \label{eq:cc}
        CC(\hat{Y},Y) = \frac{\sigma(\hat{Y},Y)}{\sigma(Y) \times \sigma(\hat{Y})}
    \end{equation}
    
\end{itemize}

\section{\uppercase{{Experiment Result}}}

In this section, we first present results of both baseline and our proposed approach using part of our pre-processed dataset (GazeMining and Contrastive Website), as outlined in Section~\ref{sec:meth}. Then, we provide a comparison of best results of our approach with alternative saliency prediction models using full set of our pre-processed dataset, establishing a benchmark for eye-gaze prediction task given webpage screenshot inputs.



\subsection{Impact of Fine-tuning using independent and combined dataset.}

\begin{table}[]

\setlength\tabcolsep{3pt}
\centering
\resizebox{1\linewidth}{!}{
\begin{tabular}{c|l|ccc|ccc}
    \hline
 \multirow{2}{*}{No.}&\multirow{2}{*}{Models} & \multicolumn{3}{c|}{GazeMining}      & \multicolumn{3}{c}{Contrastive Website}    \\ \cline{3-8}
 &        & NSS   & AUC-J & CC       & NSS   & AUC-J& CC       \\ \hline
1. & P-MSI      & 1.037  & 0.699& 0.130   & 0.839 & 0.683 & 0.109\\
2. & FT-MSI-GM    & 1.398 & \textbf{0.753} & \textbf{0.170} & 0.717 & 0.651 & 0.093     \\
3. & FT-MSI-CW & 0.695& 0.655 & 0.081      & \textbf{2.364} & 0.777& \textbf{0.277} \\
4. & FT-MSI-CMB & \textbf{1.447}& 0.752& 0.170       & 2.269&\textbf{ 0.789 }& 0.269\\
     
    \hline
\end{tabular}}
\caption{Results of fine-tuned models on both preprocessed GazeMining and Contrastive Website datasets. }
\label{tab:fintune}
\end{table}

\label{subsec:finetuning}
Table~\ref{tab:fintune} shows results of four alternative models that are trained using different datasets. Here we can see that pre-trained MSI (P-MSI) model performs worse in comparison to other models, which suggests its inability to generalize to eye-gaze saliency prediction task. We can further observe that when fine-tuning is applied, improvement from original P-MSI is noticeable, especially when it is tested on similar dataset used for fine-tuning. This can be seen with higher values achieved on all metrics of FT-MSI-GM and FT-MSI-CW compared to P-MSI on GazeMining dataset and Contrastive Website dataset respectively.

When model is trained using combined dataset, however, accuracy improvement is instead lower. This is indicated by lower quantitative values achieved by FT-MSI-CMB as opposed to their counterparts (FT-MSI-GM and FT-MSI-CW), that are trained separately. This phenomenon may come from difficulty of estimator to learn from these two datasets, which are quite distinct, and also from the nature of task executed by the users on each dataset.

Figure~\ref{fig:comFT} presents visual prediction examples for baseline model (P-MSI) and best performing model (FT-MSI-CW) for Contrastive Website Dataset (where the example is originated). The screenshot example is a part of route search task, where user has to enter vehicle information to estimate fuel consumption. Thus it is natural for users to focus on dialog box (displayed in the center of the webpage screenshot) resulting in users' fixations being located within this area. 

By evaluating the predictions of eye-gaze heatmaps of baseline model of P-MSI, we can see that even though it manages to predict parts of ground-truth (eye-gaze) locations, however, it also falsely predicts other less relevant locations as eye-gaze locations (i.e. advertisement part). These results come from its tendency to detect high color contrast as area of interest, which is common in natural image datasets  (SALICON~\cite{jiang2015salicon}). However, this leads to higher occurrences of false positive, thus reducing its accuracy. Our fine-tuned model however, manages to reduce existing inaccuracies by absorbing the characteristics of dataset during fine-tuning, adapting model to this specific eye-gaze prediction task. In this example, we see that the predictions of fine-tuned models are more precise, especially on dialog box and they exclude some of high contrast area that are normally recognized as saliency area (e.g. advertisement locations). 

From the experimental result, it can be concluded that saliency detection model trained on natural image dataset does not work well for direct application to web screenshot scenario. This can be mitigated by performing careful fine-tuning to only use a dedicated dataset. Given this result, then we use FT-MSI-GM and FT-MSI-CW as base for MMIAN optimization, and for comparison in the next section.

\subsection{Impact of Multi-modal Masks and Attention-based Fusion}
\begin{figure*}[h!]
\centering
    \includegraphics[width=1\linewidth]{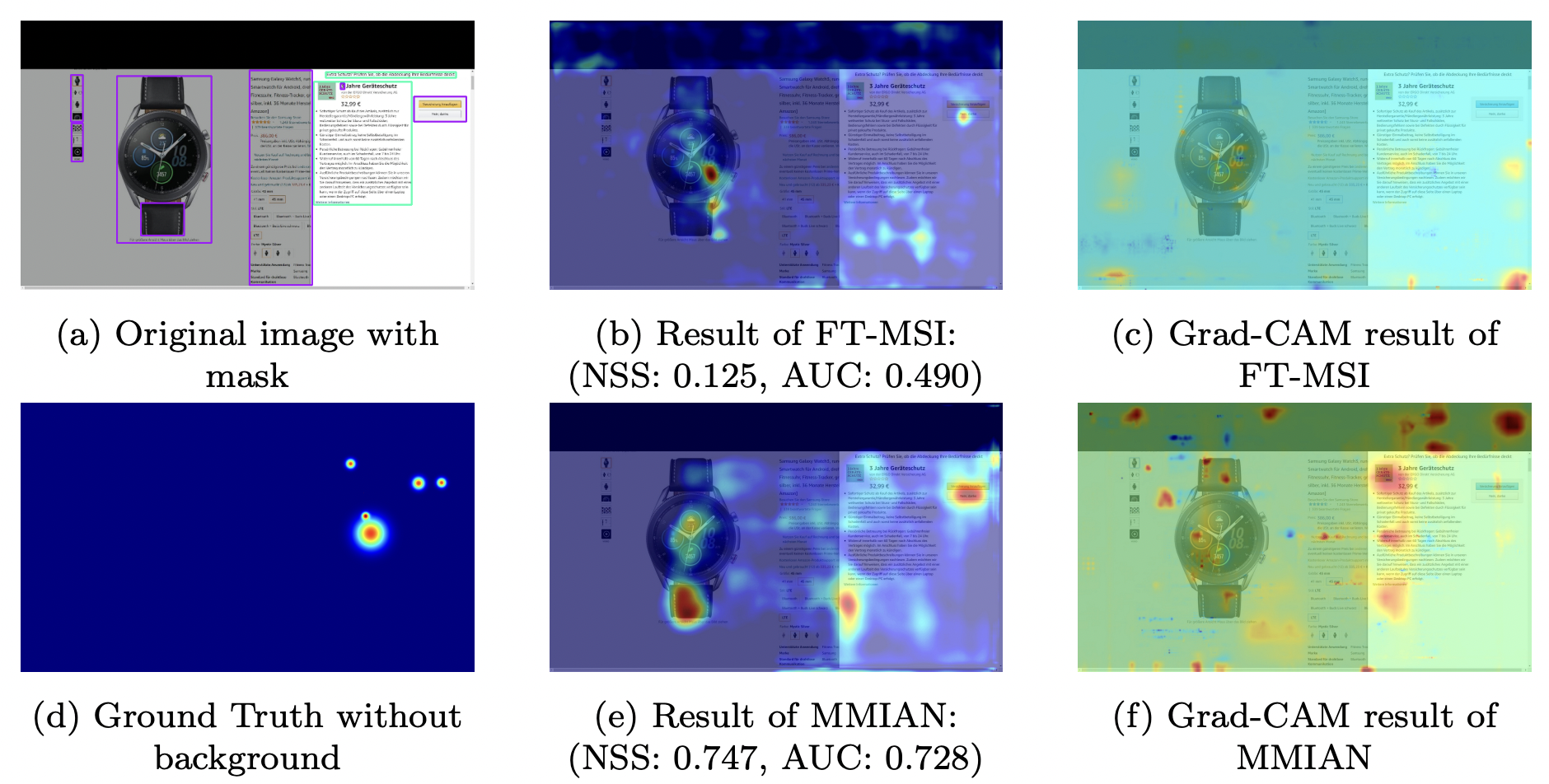}
    \caption{Example of prediction results of MMIAN and Fine-tuned MSI Model (FT-MSI). First column of (a) and (d) shows webpage screenshot with outlined text (light green box) and image (purple box) area, along with eye-gaze heatmap ground truth. Second column (b) and (e) shows predicted result from FT-MSI and MMIAN model, with third column of (c) and (f) shows respective Grad-CAM heatmap.
    }
    \label{fig:gradcam}
\end{figure*}

Table~\ref{tab:aff} presents results of our proposed MMIAN model and best results of fine-tuned MSI-Model from previous section. From this result, it can be seen that MMIAN produces higher value across quantitative metrics in overall, outperforming results from fine-tuned MSI-Model. The gains on all of these metrics altogether, suggest that estimation results of MMIAN are more accurate compared to the best of FT-MSI results (from FT-MSI-GM and FT-MSI-CW), with improvements on both true positive (as judged by AUC-Judd) and false positive Instances (as evaluated by NSS and CC score). 



In order to provide a more comprehensive analysis of the impact of the use of image and text Masks and attention mechanisms, we first show locations of detected images and texts on webpage screenshot input to provide semantic explanations of their relevances for our MMIAN model to produce accurate predictions. Then, we implemented a Gradient-weighted Class Activation Mapping (Grad-CAM)~\cite{selvaraju2017grad} by propagating multiplied error value and gradient to each convolutional layer, enabling us to investigate the relevant activation of convolutional kernels - with respect to image input - indicating most prominent part of webpage screenshot for prediction. The example of original webpage screenshot input that includes image and text locations (drawn as bounding box), associated eye-gaze location ground-truth, generated Grad-CAM and each prediction of fine-tuned MSI and MMIAN is shown in Figure~\ref{fig:gradcam}. 

On Figure~\ref{fig:gradcam}, first we can see that prediction of MMIAN is more accurate than FT-MSI, with larger and correctly identified eye-gaze areas are present. This is especially noticeable in areas where both image and text are present. For examples there are areas where button resides (that is recognized as image) and product description, which in this case, are indeed the locations where user looks at. This perceived higher accuracy can be attributed to the use of image and text masks during learning, which is combined effectively through attention mechanism to properly 'guide' learning process to put priority in this area. This impact is also apparent from observation of grad-cam heatmap of FT-MSI and MMIAN, where grad-cam heatmap of MMIAN are seen to be more concentrated on both of images and text area simultaneously (indicating where models' attention is) as opposed to ones produced by FT-MSI. Lastly, we also notice that there are more frequent activations of grad-cam heatmaps of MMIAN compared to ones from FT-MSI, which may suggest that larger perceptual field of view from MMIAN is indicative to be able to produce more accurate eye-gaze estimations in overall.

\subsection{State-of-the-Arts Comparison}

We compare  best predictions of our approach against other available, seven generalized visual saliency models:

\begin{enumerate}
    \item Context-Aware Saliency Detection (CASD)~\cite{goferman2011context}.
    \item Discrete Cosine Transform (DCTS)~\cite{houx2012image}.
    \item Hypercomplex Fourier Transform (HFT)~\cite{li2012visual}.
    \item Incremental Coding Length (ICL)~\cite{hou2008dynamic}.
    \item RARE~\cite{riche2012rare}.
    \item SeoMilanfar~\cite{seo2009static}.
    \item Spectral Residual (SR)~\cite{hou2007saliency}.
\end{enumerate}
including one specialized eye-gaze location estimator of Multiple Kernel Learning (MKL)~\cite{shen2014webpage}. 

Table~\ref{tab:ext-compare} shows results from all evaluated approaches. Here we see state-of-the-art results of MMIAN that outperforms other alternatives, including eye-gaze predictor of MKL with a large margin on FiWI dataset (note that there is no training involved for FiWI dataset evaluation). This result is mainly due to large differences in task characteristics between general visual saliency prediction (where first seven models are trained) and webpage screenshot based eye-gaze estimations tasks (where MKL and MMIAN are specialized). This highlights the inability of visual saliency based models to generalize on this specific task. Furthermore, our approach performs better than MKL on FiWI, demonstrating the effectiveness of our overall approach for eye-gaze estimation task. 

\begin{table}[h!]

\setlength\tabcolsep{3pt}
\centering
\resizebox{1\linewidth}{!}{
\begin{tabular}{c|l|ccc|ccc}
    \hline
\multirow{2}{*}{No.}&\multirow{2}{*}{Models} & \multicolumn{3}{c|}{GazeMining}& \multicolumn{3}{c}{Contrastive Website}       \\ \cline{3-8}
&      & NSS   & AUC-J & CC       & NSS   & AUC-J & CC     \\ \hline
1. & FT-MSI      & 1.398& 0.753&  0.170  & 2.364 &0.777 &0.277 \\ 
2. & MMIAN      & \textbf{1.579}& \textbf{0.764}& \textbf{0.184}       & \textbf{2.487}  & \textbf{0.787} & \textbf{0.292}\\
    \hline
\end{tabular}}
\caption{Quantitative results of both fine-tuned MSI models and MMIAN model.}
\label{tab:aff}
\end{table}

\begin{table*}[h]
\centering
\resizebox{\textwidth}{!}{%
\begin{tabular}{c|l|ccc|ccc|ccc} \hline
 \multirow{2}{*}{No.}&\multirow{2}{*}{Models} & \multicolumn{3}{c|}{GazeMining}      & \multicolumn{3}{c|}{Contrastive Website} & \multicolumn{3}{c}{FiWI}    \\ \cline{3-11}
 &   & \multicolumn{1}{c}{NSS} & \multicolumn{1}{c}{AUC} & \multicolumn{1}{c|}{CC} & \multicolumn{1}{c}{NSS} & \multicolumn{1}{c}{AUC}& \multicolumn{1}{c|}{CC}  & \multicolumn{1}{c}{NSS} & \multicolumn{1}{c}{AUC} & \multicolumn{1}{c}{CC} \\ \hline
1.& CASD~\cite{goferman2011context}       & 0.567& \textcolor{blue}{0.653}& 0.064 & 0.419& \textcolor{red}{0.614}& 0.053 & 0.680& 0.732& 0.233\\
2.& DCTS~\cite{houx2012image}      & 0.479& 0.618& 0.053 & 0.256& 0.552& 0.035 & 0.541& 0.671& 0.195\\
3.& HFT~\cite{li2012visual}       & \textcolor{red}{0.707}& 0.644 & \textcolor{red} {0.088}& \textcolor{red}{0.534}& 0.593& \textcolor{red}{0.067} & 0.740& 0.737 & 0.251\\
4.& ICL~\cite{hou2008dynamic}      & 0.485& 0.518& 0.057 & 0.192& 0.490& 0.030 & 0.444& 0.618& 0.162\\
5.& RARE~\cite{riche2012rare}     & \textcolor{blue}{0.632}& \textcolor{red}{0.653}& \textcolor{blue}{0.072} & 0.382& 0.589 & 0.052& \textcolor{blue}{0.850}& \textcolor{blue}{0.758}& \textcolor{blue}{0.280}\\
6.& SeoMilanfar~\cite{seo2009static}& 0.393& 0.584& 0.038 & 0.350& 0.571& 0.044 & 0.445& 0.651& 0.163\\
7.& SR~\cite{hou2007saliency}       & 0.566& 0.639& 0.062 & \textcolor{blue}{0.510}& \textcolor{blue}{0.612}& \textcolor{blue}{0.062} & 0.635& 0.714& 0.216\\
8. & MKL~\cite{shen2014webpage}    & \textbf{-}   & \textbf{-}    & \textbf{-}     & -   & -    & -     & \textcolor{red}{1.200}& \textcolor{red}{0.702}& \textcolor{red}{0.382}\\
9. & \textbf{MMIAN (proposed)}       & \textbf{1.579}& \textbf{0.764}& \textbf{0.184}  & \textbf{2.487}& \textbf{0.787} & \textbf{0.292} & \textbf{1.385}& \textbf{0.786} & \textbf{0.397}\\ 


\hline
\end{tabular}%
}
\caption{Comparison of existing saliency predictors evaluated on test sets of our full pre-processed dataset. The boldface indicates best results, red color implies second-best results, and third-best results are marked by blue coloured fonts.}
\label{tab:ext-compare}
\end{table*}

\begin{figure*}[]
    \centering
    \includegraphics[width=\linewidth]{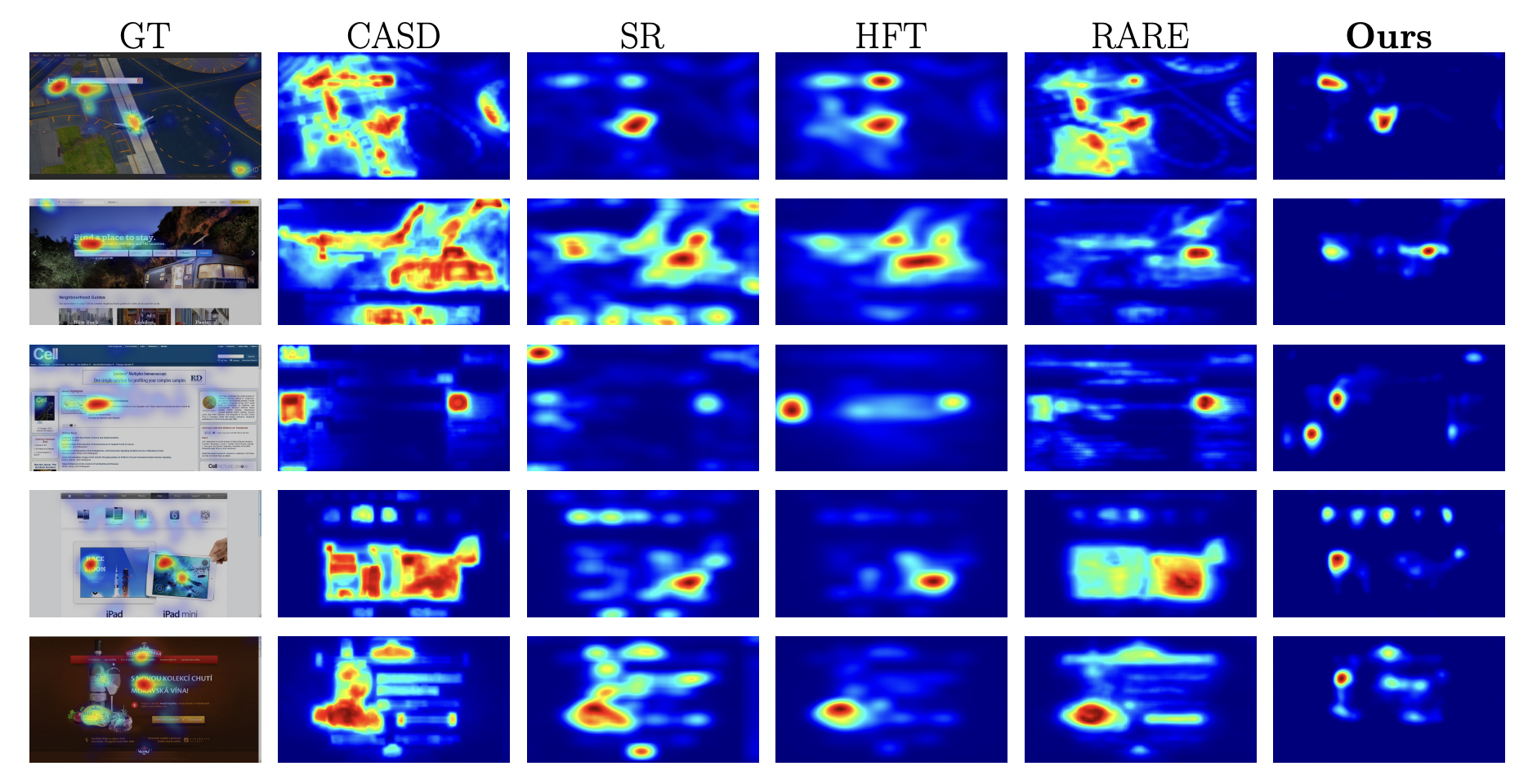}
    \caption{Five examples of screenshot inputs from FiWI dataset (with respective eye-gaze heatmaps ground-truth overlaid) and predictions from other five saliency predictors (including ours).}
    \label{fig:extCompare}
\end{figure*}

Figure~\ref{fig:extCompare} shows predictions example of our MMIAN and four other best performing model on FiWI for evaluation (excluding MKL, due to lack of available implementation of the model). Based on this figure, we can observe that most of alternative models produce a large area of webpage screenshot as potential eye-gaze locations. However, this leads to largely false positive prediction, given the mismatch between predictions against actual eye-gaze locations from user of these webpage screenshots (i.e. most area are falsely identified as eye-gaze locations). In contrast, our approach produces more precise estimates, as observed by more refined (and accurate) predicted area of eye-gaze locations (that explains large margin in terms of NSS metrics value from our models compared to others, as shown in Table~\ref{tab:ext-compare}). This is mainly due to the tendency of visual saliency models to focus on pure appearance of webpage screenshot (i.e. high contrast images), as opposed to MMIAN that has been conditioned to the specific characteristics of our pre-processed eye-gaze dataset (which inherently contains eye-gaze characteristics of users, given webpage screenshot). Remarkably, our models is also capable of correctly predicting eye-gaze locations on relevant areas, such as text (third row), image (fourth row) and their combinations (first, second and fifth row), in comparison with other approaches. Here we see that our models' predictions are consistently more accurate than alternatives, demonstrating the effectiveness of our approach.

\section{\uppercase{{Conclusion}}}






In this work, we enable the development of automatic eye-gaze estimations given webpage screenshot inputs to enable the improvement of the webpage layout, and hence respective user interaction. We do this by developing a unified eye-gaze dataset from three available website and user interaction-based datasets: GazeMining, Contrastive Website and FiWI. We then pre-process each dataset to produce necessary data for eye-gaze prediction task, such as webpage screenshot and corresponding eye-gaze heatmaps as ground truth. In addition, we generate image and textual locations from webpage (in the form of masks) which can be used for training and modeling. Given our unified eye-gaze dataset, then we propose a novel deep learning based and multi-modal attentional network eye-gaze predictor to benefit from the characteristics of the dataset. Our proposed approach leverages spatial locations of text and image (in form of masks), which is further fused with attentional mechanisms to enhance the prediction results.

During analysis of the impact of fine-tuning using our pre-processed dataset, and the effects of training when the combined dataset is used, we found that the prediction results are indeed improved when careful fine-tuning is conducted. Then, we evaluate the prediction of our full approach (MMIAN) with respect to the ground truth, existing text and image locations from webpage screenshot, and part of Grad-CAM activations. Here we notice accurate predictions of our approach, especially on textual and image area where users are looking at, with large and wide activation of Grad-CAM that are concentrated on these locations. This observation demonstrates the benefit of the use of both image and text masks as input in combination with attention mechanism.

To measure the competitive results of our proposed approach, we compare them with other saliency prediction alternatives, establishing a benchmark for eye-gaze prediction task. In our comparison, we found state-of-the-art results of our model with high scores across quantitative metrics, and lower false positive rates than other approaches. Visual analysis further confirms our findings, that our approach produces more accurate prediction of eye-gaze locations on relevant website locations, including where text and image are present. The result suggests the superiority of our approach in capturing user behavior. Future work will be to incorporate other user behavior characteristics (e.g. mouse trajectory), including users' identity (if accessible) such as age, and locations as additional modalities to further benefit from this information to improve prediction accuracy.

\section*{\uppercase{Acknowledgement}}
This work is funded by UDeco project by Germany BMBF-KMU Innovativ - 01IS20030B.



\bibliographystyle{apalike}
{\small
\bibliography{example}}

\end{document}